\documentclass[10pt,twocolumn,letterpaper]{article}

\usepackage[pagenumbers]{cvpr} % To force page numbers, e.g. for an arXiv version

\usepackage[dvipsnames]{xcolor}

\usepackage[utf8]{inputenc} % allow utf-8 input
\usepackage{hyperref}       % hyperlinks
\usepackage{url}            % simple URL typesetting
\usepackage{booktabs}       % professional-quality tables
\usepackage{amsfonts}       % blackboard math symbols
\usepackage{nicefrac}       % compact symbols for 1/2, etc.
\usepackage{microtype}      % microtypography
\usepackage{enumitem}
\usepackage{graphicx}
\usepackage{multirow}
\usepackage{pifont}
\usepackage[dvipsnames]{xcolor}
\usepackage{tcolorbox}
\usepackage{adjustbox}
\usepackage{natbib}
\usepackage{colortbl}      % 引入colortbl包以使用\cellcolor

\usepackage{cinzel}
\usepackage{helvet}
\usepackage{xspace}

\newcommand{\dataname}[0]{\textsc{AssistGUI}\xspace}

\usepackage[T1]{fontenc} 
\newcommand{\modelname}[0]{\textcinzel{\textit{ACE}}}
\usepackage{mdframed}

\definecolor{mypurple}{RGB}{200,192,248}
\definecolor{mypurpledeep}{RGB}{142,126,240}
\definecolor{mygreen}{RGB}{117,170,156}
\definecolor{myyellow}{RGB}{255,192,0}
\definecolor{myblue}{RGB}{57,143,255}
\definecolor{mygrey}{RGB}{231,230,230}
\definecolor{codey}{RGB}{220,220,170}
\definecolor{coder}{RGB}{206,145,120}
\definecolor{codeb}{RGB}{156,220,254}
\definecolor{codenum}{RGB}{204,204,204}

\title{\includegraphics[scale=0.42, bb=0 15 50 34]{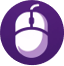} \dataname: Task-Oriented Desktop Graphical User Interface Automation}

\definecolor{pearThree}{HTML}{E74C3C}
\definecolor{pearDark}{HTML}{2980B9}
\definecolor{pearDarker}{HTML}{1D2DEC}

\usepackage[dvipsnames]{xcolor}

\hypersetup{
 colorlinks,
 citecolor=pearDark,
 linkcolor=pearThree,
    breaklinks=true,
 urlcolor=pearDarker}

\def\paperID{2639} % *** Enter the Paper ID here
\def\confName{CVPR}
\def\confYear{2024}

\author{Difei Gao, Lei Ji, Zechen Bai, Mingyu Ouyang, Peiran Li, Dongxing Mao, Qinchen Wu, \\ Weichen Zhang,  Peiyi Wang, Xiangwu Guo, Hengxu Wang, Luowei Zhou, Mike Zheng Shou$^{\dag}$\\
    \small $^{\dag}$Corresponding author \\ Show Lab, National University of Singapore \\
\small \url{https://showlab.github.io/assistgui/}
}

\begin{document}
% \maketitle
\twocolumn[{%
\renewcommand\twocolumn[1][]{#1}%
\maketitle
\begin{center}
    \centering
    \captionsetup{type=figure}
    \vspace{-0.8cm}
    \includegraphics[width=0.88\linewidth]{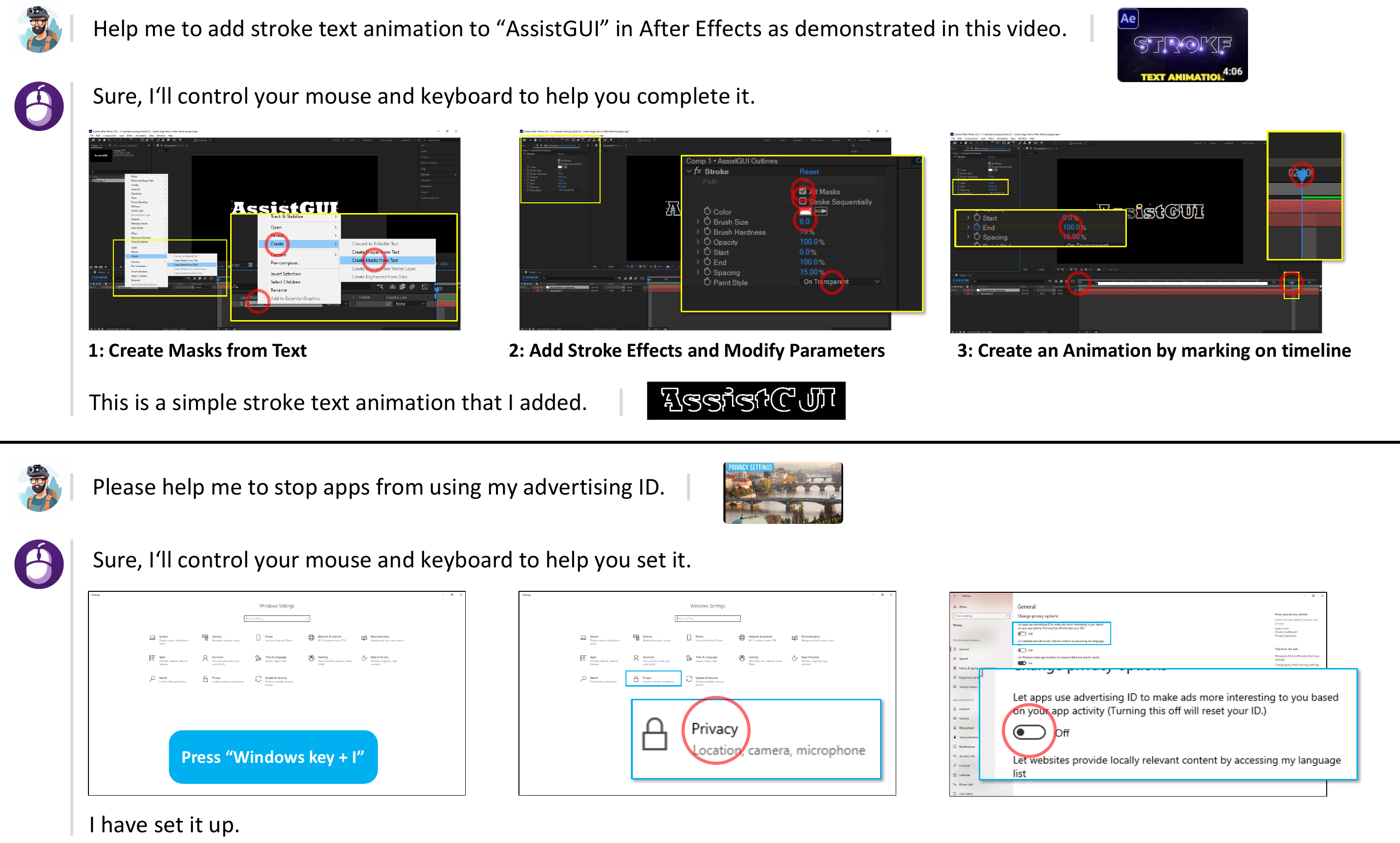}
    \vspace{-0.2cm}
    \captionof{figure}{\textbf{Illustration of GUI Task Automation in AssistGUI.} Given a user query and an instructional video for reference, an agent is required to manipulate the keyboard and mouse on the desktop to complete the task as requested by the user. }
\label{fig:teaser}
\end{center}%
}]
\begin{abstract}
Graphical User Interface (GUI) automation holds significant promise for assisting users with complex tasks, thereby boosting human productivity.  Existing works leveraging Large Language Model (LLM) or LLM-based AI agents have shown capabilities in automating tasks on Android and Web platforms. However, these tasks are primarily aimed at simple device usage and entertainment operations. This paper presents a novel benchmark, \dataname, to evaluate whether models are capable of manipulating the mouse and keyboard on the Windows platform in response to user-requested tasks. We carefully collected a set of 100 tasks from nine widely-used software applications, such as, After Effects and MS Word, each accompanied by the necessary project files for better evaluation. Moreover, we propose an advanced Actor-Critic Embodied Agent framework, which incorporates a sophisticated GUI parser driven by an LLM-agent and an enhanced reasoning mechanism adept at handling lengthy procedural tasks. Our experimental results reveal that our GUI Parser and Reasoning mechanism outshine existing methods in performance. Nevertheless, the potential remains substantial, with the best model attaining only a 46\% success rate on our benchmark. We conclude with a thorough analysis of the current methods' limitations, setting the stage for future breakthroughs in this domain. 
\end{abstract}    
\section{Introduction}
\label{sec:intro}

Novices often face a steep learning curve when acquainting themselves with complex desktop applications. For instance, software like After Effects and Premiere Pro offer a suite of advanced functions for video editing, yet its richness in features sets a high entry barrier for new users. An AI Assistant with the capacity to comprehend GUI interfaces, grasp software usage methodologies, and manipulate applications would significantly expedite the learning and operating processes. As such an Assistant evolves, it will liberate humans from the tedious complexities that currently impede their creativity and productivity.

Early software automation methods, exemplified by voice assistants such as Siri or Alexa, rely on predefined intents and the extraction of parameters from user queries to execute functions, lacking the flexibility required for complex operations. With the advent of generative models, e.g., GPT~\cite{openai2022chatgpt}, there has been a shift towards using Large Language Models (LLMs)~\cite{llama, touvron2023llama, alpaca} or LLM agents~\cite{yao2022react, yang2023mm} to formulate interactive tasks as a text-to-text generation. 
Several benchmarks~\cite{liu2023agentbench} are proposed to evaluate their performances on using an Ubuntu bash terminal, using a database, or engaging in card games, with some recent works~\cite{openai2023gpt4, openai2022chatgpt} demonstrating impressive results.
Moreover, some benchmarks~\cite{rawles2023android,wen2023empowering,yao2022webshop,shi2017world} are proposed to evaluate Web navigation and Smartphone manipulation. Some work has proposed methods based on HTML~\cite{wen2023empowering,rawles2023android} and pure vision~\cite{shaw2023pixels,yan2023gpt}. ~\cite{yan2023gpt} utilized GPT-4V-SoM~\cite{yang2023setofmark} for Smartphone GUI Navigation, which has achieved promising results. While these studies are indeed exciting, these tasks are primarily centered around entertainment scenarios. Consequently, an agent's proficiency in these tasks may not necessarily lead to a substantial increase in human productivity.

Therefore, this paper aims to evaluate the model on task-oriented Desktop Graphical User Interface Automation, aimed at assessing model performance in utilizing productivity software. This task poses unique challenges compared to previous Web and Android Automation:
\begin{itemize}%[leftmargin=0.2cm, itemindent=0.3cm]
  \item \emph{Dense GUI Understanding}: This involves interpreting various forms of information, not only salient texts on the screen but also various visual elements like icons and footage in the office or design software.
  \item \emph{Complex Operations}: Desktop operations demand more sophisticated actions than those on the Web or Smartphone, extending beyond basic tapping, typing, etc. to include operations like dragging files or drawing masks on footage.
  \item \emph{Long Procedure}: Executing a task in productivity software can involve a sequence of complex steps. For example, creating a single effect in AE will include layer creation, media import, effect adding, animation creation, etc.
\end{itemize}
In order to better research this important but still largely unexplored domain, we introduce \dataname, a benchmark designed for Desktop GUI Automation. As illustrated in Figure~\ref{fig:teaser}, the model receives an instructional video demonstrating a specific function of an application, along with a user query pertinent to the video's content. The model's objective is to interact with the software to fulfill the task specified in the query. The inclusion of instructional videos is crucial, particularly for tools like After Effects, which have a vast array of user-developed customized features. This design aims to make the model adaptable and efficient at acquiring new usage techniques. 

Correspondingly, we constructed a benchmark that spans 5 major categories of desktop tasks: office work, design, widget usage, system setting, and file manipulation, covering 9 popular applications, such as Premiere Pro, After Effect, PowerPoint, etc. In total of 100 specific tasks are provided, each accompanied by a textual query, an instructional video, and carefully created project files. In addition to the data, we have developed a system that enables a local Windows environment to be presented as an interactive platform to a remote server, facilitating model development and testing.

In addition, we introduce a robust baseline, an embodied agent framework, named Actor-Critic Embodied Agent \modelname, as depicted in Figure~\ref{fig:overview}. Specifically, drawing from the concept of LLM-based Agents~\cite{yang2023mm,shen2023hugginggpt,gao2023assistgpt}, we develop an advanced GUI parser that can identify a variety of UI elements. Moreover, we propose a novel reasoning approach that allows for the hierarchical decomposition of tasks and dynamically adjusts future steps by evaluating the results of each step, sharing the spirit of Actor-Critic algorithm~\cite{konda1999actor}. Our experiments on the \dataname benchmark revealed that while the proposed model demonstrates promising potential, it also underscores the task's inherent complexity. Subsequent ablation analysis of different components within our agent framework revealed limitations in both current LLMs, LMMs, and LLM-based agents when it comes to intricate GUI automation tasks. These insights lead us to suggest future directions for improvement in GUI understanding and action generation for desktop GUI applications.

In summary, our work makes the following contributions:
\begin{itemize}%[leftmargin=0.2cm, itemindent=0.3cm]
    \item We introduce, to the best of our knowledge, the first task specifically designed for desktop software automation.
    \item We have created a comprehensive benchmark featuring a carefully selected collection of samples and developed environments that aid in evaluation.
    \item We present a strong baseline equipped with advanced GUI perception capability and a new planning mechanism.
    \item Extensive experimentation assesses our approach's effectiveness and highlights the challenges in desktop GUI automation for existing models.
\end{itemize}

\section{Related Work}
\label{sec:related_work}

\textbf{UI Task Automation Benchmark.}
UI automation tasks mainly focus on mobile or web applications with both environment development and benchmark construction. The \emph{mobile} scenarios are widely studied with open-source environments built on top of the Android ecosystem. The environments \cite{toyama2021androidenv,schneider2022mobile} provide an interactive way for reinforcement learning for relatively simple tasks.  
The benchmarks \cite{li2020mapping,burns2022dataset,rawles2023android,wen2023empowering} further extend to more diverse and complex low-level or high-level tasks.

Additionally, there are several simulated \emph{web} environments developed for agents to learn in an interactive way\cite{shi2017world,pasupat2018mapping,yao2022webshop,deng2023mind2web,zhou2023webarena}. 
Regarding further computer tasks, NL2Bash\cite{lin2018nl2bash} and agentbench\cite{liu2023agentbench} provide interaction with the \emph{terminal} systems taking language as inputs and outputs. Different from them, ours is more challenging to handle graphical interaction within a real-world Desktop environment for complex UI and diverse tasks.

% \noindent 
\textbf{LLM-as-Agent.}
Recent studies present promising research directions prompting LLM for multi-step reasoning and invoking application-specific APIs, external tools, or domain-specific models.
Some works~\cite{wei2022chain,yao2022react,schick2023toolformer, yao2022react,shinn2023reflexion,schick2023toolformer,paranjape2023art}, such as CoT, and ReAct, enhance the model's capability for better conversation by logical reasoning. There is also a growing body of work~\cite{yang2023mm,suris2023vipergpt,wu2023visual,shen2023hugginggpt,lu2023chameleon,gao2023assistgpt} focusing on using LLMs in conjunction with visual tools to perform multimodal tasks, such as visual question answering and video summarization. Some research~\cite{wu2023visual} even proposes LLM-based agents for image editing. We introduce a specialized LMM-based agent tailored for Desktop GUI Automation, aiming to provide a powerful baseline for this task.

\textbf{Embodied AI for UI Task Automation.}
The significant challenges of GUI task automation are the understanding of the complex graphical UI observation and the planning to achieve various tasks, leading to \emph{end-to-end} supervised approaches or LLM-based zero-shot \emph{two-stage} solutions. Previous end-to-end methods adopt reinforcement learning\cite{gur2018learning} or imitation learning\cite{humphreys2022data}. \cite{sun2022meta,zhan2023you,shaw2023pixels,brohan2023rt,zhang2023reinforced} rely on vision-language-action pretraining to learn to directly map visual observation to actions. However, these methods usually require a significant number of human expert demonstrations, which are still hard to generalize to the general applications. With the advent of LLM, there are some LLM-based two-stage methods. The first stage is to semantically understand the elements of the observed UI by either off-the-shelf models like OCR or learnable vision-language models  \cite{lee2023pix2struct,banerjee2023lexi,he2021actionbert,bai2021uibert}. 
For example, \cite{wen2023droidbot,wen2023empowering,yan2023gpt} propose to convert GUI into HTML representation or natural language sentence.
Consequently, the second stage is to generate executable steps given the UI elements \cite{yao2022webshop,zhou2023webarena,kim2023language,li2023zero} usually with LLM. However, single OCR and vision-language models are limited to simple GUIs and fail to capture the full complexity of Desktop GUIs. They also struggle with long processes due to their single-step generation approach. To address these limitations, we've developed an LLM-based agent equipped with diverse tools for parsing various UI elements and a new hierarchical planning and critic mechanism for handling extended procedures.

\section{\dataname Benchmark}
\label{sec:benchmark}
\dataname benchmark provides real-world interactive environment, dataset across broad tasks and goal-oriented evaluation.

\subsection{Task Formulation}

Desktop task automation in \dataname can be formulated as follows: given a natural language query that briefly describes a specific task, along with an instructional video as a supplement that more detailed illustrates how to complete it, and the relevant application, the output is a sequence of UI actions to fulfill the user's query. 

\textbf{Task description.} To describe the task, a textual request $\boldsymbol{q}$ is provided by the user, which describes the functionality of an application to be accomplished, e.g., \emph{Center align the text "AssistGUI" in my opened After Effect project}. For some functions of productivity tools, there might be multiple user-developed implementations. We aim for the model to generate actions based on the given references. Thus, an instructional video, denoted as $\boldsymbol{v}$, is also provided. 

\textbf{State observation.} The state of the environment is composed of two types of information. The first type stems from the operating system's textual metadata about the software being used. In contrast to web pages, where HTML offers comprehensive information, much of this metadata in desktop applications is internal and thus not readily accessible. As a result, the metadata mainly includes the layout of panels and pop-up windows. The second type of information consists of screen captures, which offer a more holistic view by providing visual context.

\begin{figure}[tbp]
\centering
\includegraphics[width=0.45\textwidth]{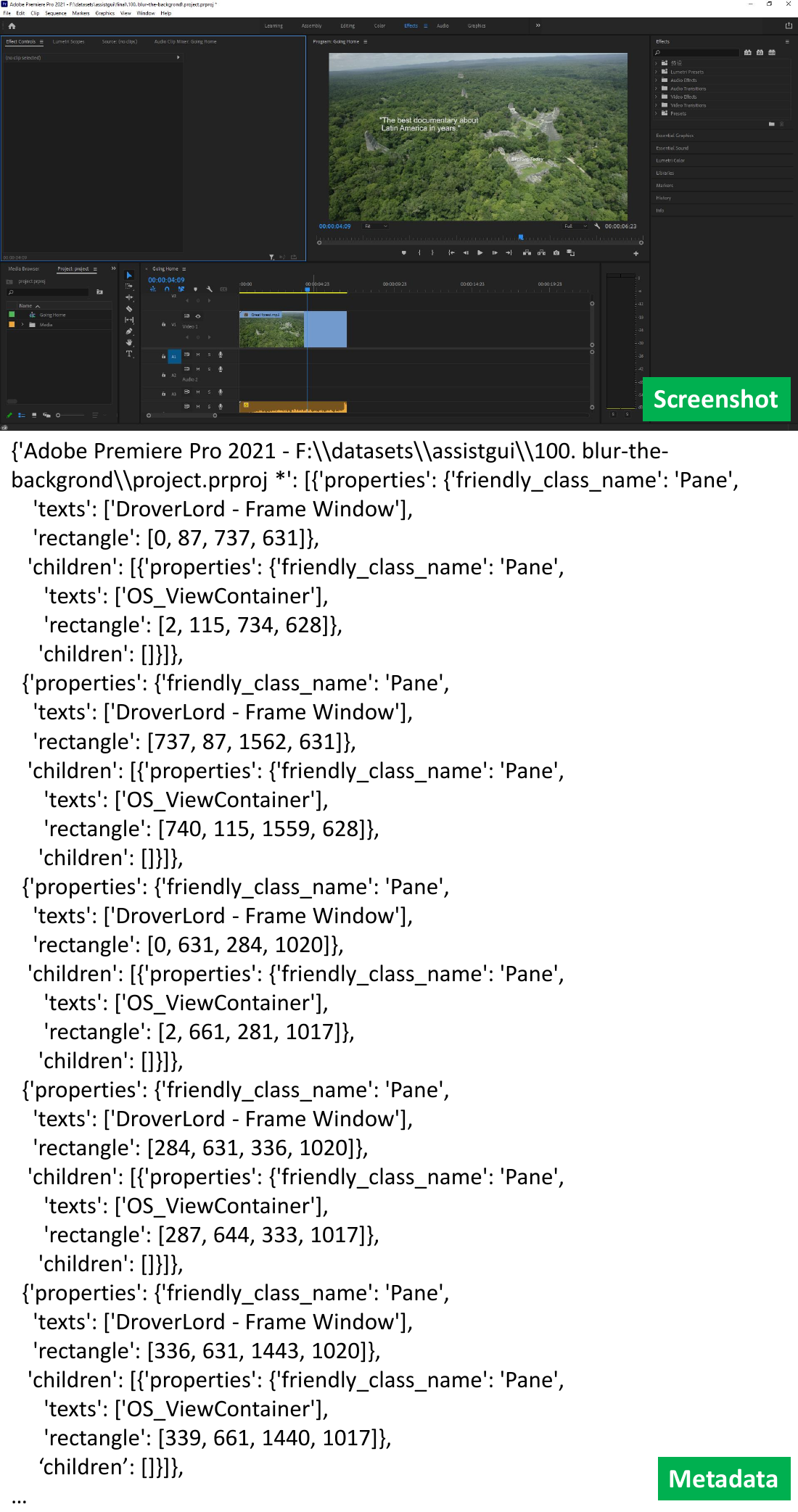} % Include the image
% \vspace{-0.3cm}
\caption{One example of screenshot and metadata.} % Caption for the figure
\label{sup_fig:fig1} % Label for referencing the figure in the text
% \vspace{-0.3cm}
\end{figure}

In Figure~\ref{sup_fig:fig1}, we present an example of the metadata and screenshot. It's worth mentioning that for software like Premier Pro, it is challenging to obtain meta-data that encompasses all information of the software. The main information obtainable is about large panels, while specific texts and buttons are almost impossible to extract from the meta-data. Therefore, the model must rely on visual perception capabilities to process screenshots.

\textbf{Action space.} Our action space consists of all the raw mouse and keyboard actions, including left-click, right-click, double-click, drag, keystrokes, and combinations of keys for shortcuts, among others. Mouse-related operations also include the target position at the pixel space of the observed screenshot. To construct a universal and complete representation of actions, we exactly followed a widely utilized Python library for controlling the mouse and keyboard, PyAutoGUI. One action is denoted by the syntax \texttt{action\_type(arguments)}. Here are some examples of actions that are supported in \dataname:

% \texttt{dragTo(100, 100)}, which indicates the execution of a drag action from the current position to the coordinate (100, 100).
\begin{itemize}[leftmargin=13pt]
    \item \textbf{Mouse Movement:}
    Move the mouse cursor to a specific position on the screen.
    
    \textit{Example:} \texttt{moveTo(100, 150)}

    \item \textbf{Mouse Clicks:}
    Automate mouse clicks at a specified location.
    
    \textit{Example:} \texttt{click(200, 220)}

    \item \textbf{Typing and Sending Keystrokes:}
    Simulate typing text or pressing keys.
    
    \textit{Example:} \texttt{write('Hello, world!')}

    \item \textbf{Keyboard Hotkey Combinations:}
    Press and release keyboard shortcuts or hotkeys.
   
    \textit{Example:} \texttt{hotkey('ctrl', 'c')}

    \item \textbf{Scrolling the Mouse:}
    Automate mouse scrolling up or down.
    
    \textit{Example:} \texttt{scroll(-200)} for scrolling down.

    \item \textbf{Drag and Drop:}
    Automate drag and drop actions.
    
    \textit{Example:} \texttt{dragTo(100, 200, duration=2)}

    \item \textbf{Mouse Down and Mouse Up:}
    Hold down and release the mouse button.
    
    \textit{Examples:} \texttt{mouseDown(); mouseUp()}

    \item \textbf{Press and Release Keys:}
    Press and release individual keyboard keys.
    
    \textit{Examples:} \texttt{press('enter')}

    \item \textbf{Key Down and Key Up:}
    Hold down and release a keyboard key.
    
    \textit{Examples:} \texttt{keyDown('shift')}

\end{itemize}

\textbf{Environment Implementation.}
Recognizing that productivity tools usually only support Windows or Mac systems, while AI models are often deployed on Linux, we've created a Python library to expose a local Windows environment as an interactive platform to a remote server. This is done using PyWinAuto API to collect metadata and screenshots from Windows. A communication system sends data to the server, and let server then sends predicted actions back to the local client for execution on the productivity tools. This setup allows remote control of the software by the server-based model through specific action commands.

\subsection{Data Collection}
Our benchmark is designed to include a broad spectrum of desktop tasks, systematically segmented into five major categories that are indicative of routine computer-based work. These categories include design, Office work (Office), system settings (Sys. Set.), widget usage (Widget), and file management (File Mani.).
The collection of task data within \dataname is achieved by the following steps:
% \begin{itemize}[leftmargin=0.4cm]

\begin{figure}[tbp]
\centering
\includegraphics[width=0.47\textwidth]{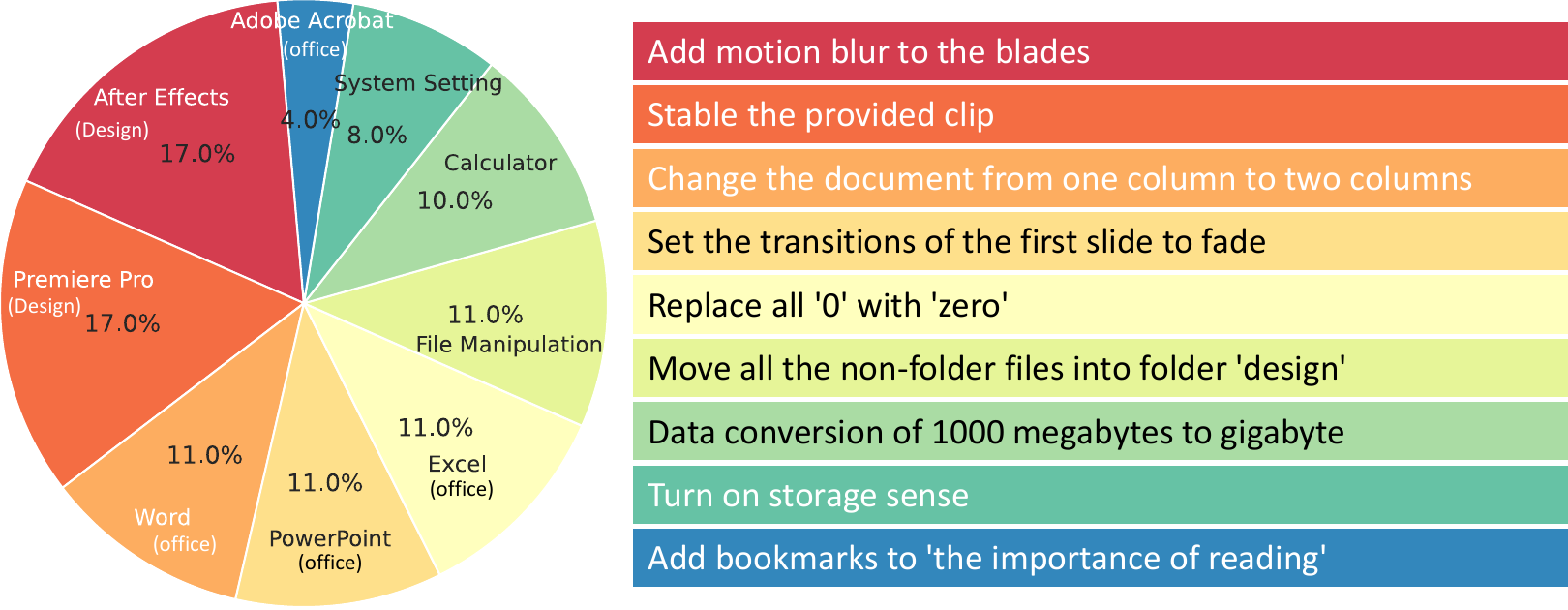} % Include the image
\vspace{-0.3cm}
\caption{\textbf{Distribution of collect tasks and one example query for each task.} We have gathered tasks across 9 applications, focusing on the use of productivity software as well as fundamental computer operations and settings.} % Caption for the figure
\label{fig:statistic} % Label for referencing the figure in the text
\vspace{-0.3cm}
\end{figure}

\begin{figure*}[t]
\centering
\includegraphics[width=\textwidth]{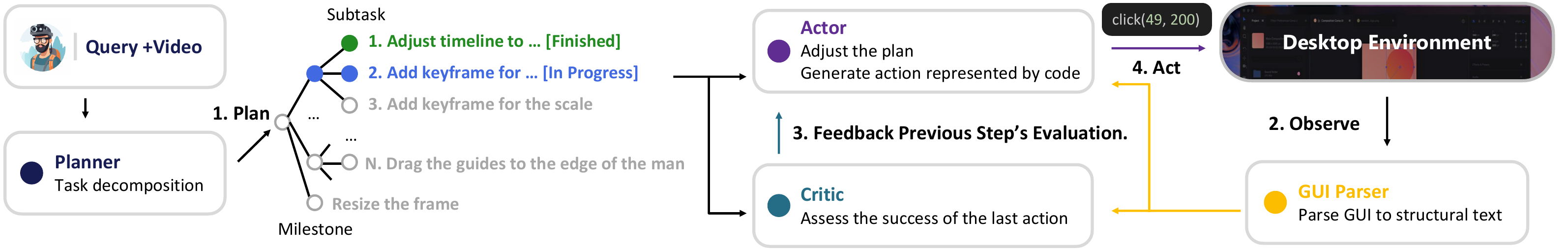} % Include the image
\vspace{-0.5cm}
\caption{\textbf{Diagram Illustration of \modelname.} It first outlines key milestones and subtasks, then iteratively employs a GUI Parser, a Critic module for action assessment, and an Actor module for adjusting the plan and generating code for controlling the desktop, sequentially completing subtasks until the task is finished.} % Caption for the figure
\label{fig:overview} % Label for referencing the figure in the text
\vspace{-8pt}
\end{figure*}

\textbf{Task Collection.}
Due to the complexity of GUI operations, at this early stage, we are primarily focusing on relatively basic tasks. We carefully select some popular instructional videos and those duration do not exceed five minutes from official software websites and video-sharing platforms.
We also manually crafted one query for each instructional video. These queries illustrate the tasks that the model is expected to complete. It is important to note that the task indicated by the query may not always align exactly with the operations shown in the video; it could include some user-customized requirements. Therefore, the model needs to modify the steps based on the instructional video, e.g., type in a different text, which aligns more closely with real-world scenarios.

\textbf{Project File Preparation.}
To make the results in the environment to be reproducible, we provide project files for all editing-related tasks. This ensures that all models initiate their tasks from an identical starting state. The project files included in our benchmark stem from two primary sources: A portion of the project files is directly sourced from the official tutorials available on the software's website. These files are typically crafted by the software providers to accompany their instructional materials. The remaining project files are meticulously prepared by annotators. We have also documented the version of each project file. The tested models are expected to modify this file using applications of the same version for fair comparison.

% \item 
\textbf{Quality Checking.} To guarantee the correctness of our benchmark, each task has undergone a quality check by letting our annotators complete the tasks within the software to verify if they yield accurate outputs. The quality check focuses on two main aspects: Firstly, it verifies the correctness of the content in the instructional video, ensuring that the demonstrated steps are accurate and lead to the anticipated outcome. Secondly, it confirms that the project files are correct and fully functional.
% \end{itemize}

\dataname finally collects 100 specific tasks from 9 commonly used applications like Premiere Pro, After Effects, and PowerPoint. We present the distribution of collected over software and show one example query for each software \dataname task in Figure~\ref{fig:statistic}.

\subsection{Evaluation} \dataname adopts an outcome-oriented evaluation approach to determine the success rate of models. Since \dataname yields several types of outputs: video output (Design), document output (Office), the final state of the software (Widget), system settings (Sys. Set.), and folder structure (File Mani.), it is hard to construct one general metric to fit all tasks, thus, we design specific metrics to calculate the success rate tailored to each type of task.

For the Design and Office tasks, we compare the similarity of the model's results with the ground truth at a pixel granularity. If it exceeds a certain threshold, it is considered successful and scores 1 point; otherwise, it scores 0. The threshold varies slightly for different tasks, depending on whether the task inherently includes a certain level of randomness. We did not adopt CLIP-Sim~\cite{wu2021godiva}, commonly used in video generation, because video editing often involves animation changes rather than semantic changes, making it difficult for CLIP to discern subtle differences. For Widget tasks, we compare the final screenshot with the ground truth, if the same in the display region (obtained by metadata), then consider it a success. For the Sys. Set. and File Mani., we write scripts to automatically determine whether the system settings and folder structure meet the expected criteria. We also standardized the specific version numbers and languages for each software.

\section{Method}
\label{sec:method}

\textbf{Overview.} 
We introduce an Actor-Critic Embodied agent, \modelname, based on LLMs that possesses the capabilities to perceive the software environment, plan actions, and execute them, as shown in Figure~\ref{fig:overview}. Specifically, the agent works in two stages: In the first stage, given a query and a video, the Planner creates a high-level plan outlining the key milestones and subtasks of the task. The second stage involves the collaborative work of three modules to sequentially accomplish these subtasks. The GUI Parser observes the GUI environment, the Critic module assesses the quality of the previous action, and the Actor then adjusts the plan based on this assessment and generates code to control the desktop.

\subsection{Planner}
\label{sec:planner}

\begin{figure}[tbp]
\centering
\includegraphics[width=0.45\textwidth]{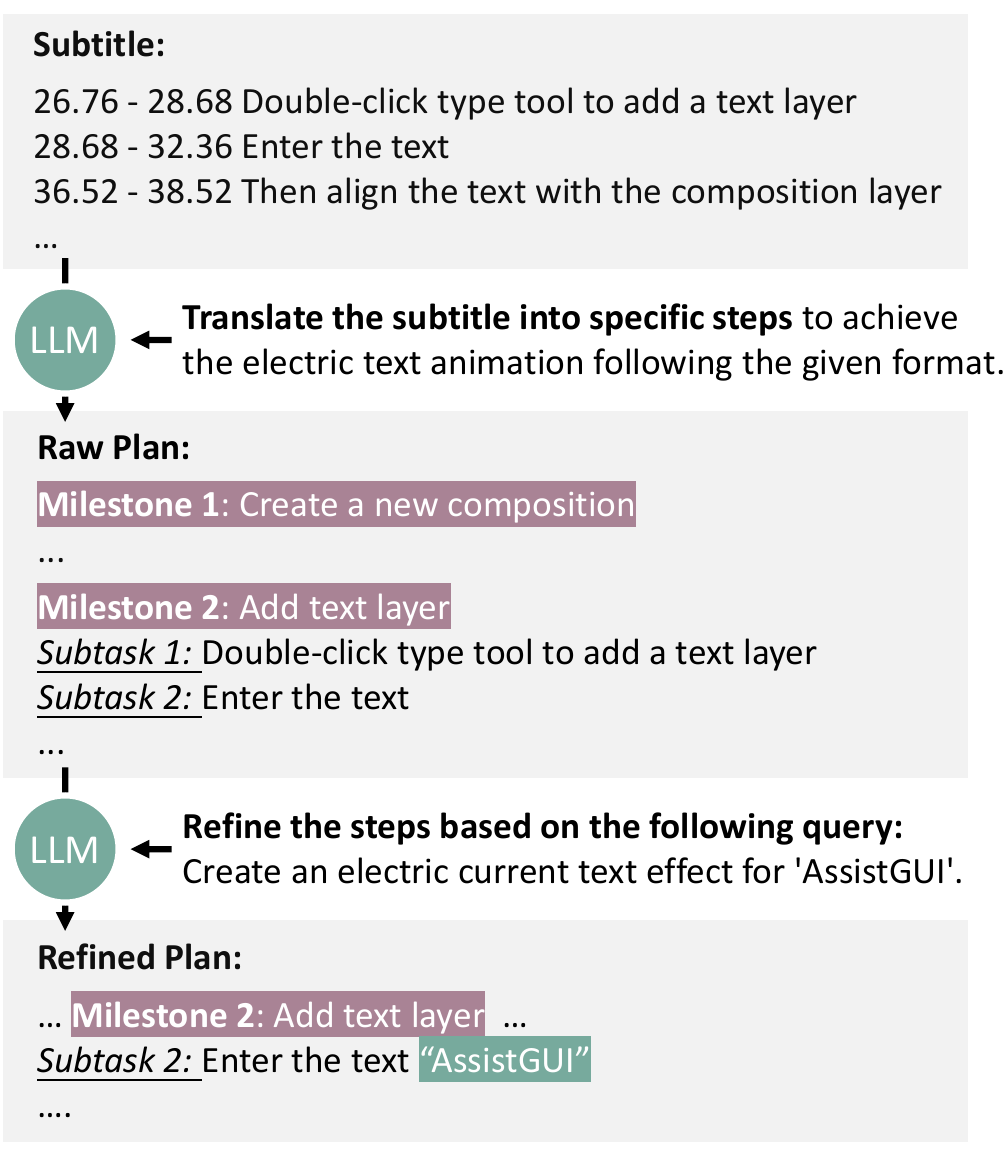} % Include the image
\caption{\textbf{Diagram Illustration of Planner.} The Planner first translates video subtitles into a structured raw plan with milestones and subtasks. It then refines this plan by specifying the user-provided query.} % Caption for the figure
\label{sup_fig:plan} % Label for referencing the figure in the text
\end{figure}

For a given query $\boldsymbol{q}$ and instructional video $\boldsymbol{v}$, the Planner aims to output a hierarchical task tree $\boldsymbol{p} = [\boldsymbol{p}_1, \boldsymbol{p}_2, ..., \boldsymbol{p}_N]$, where $\boldsymbol{p}_i$ is a text string describe the $i$-th milestone of the task. And each $\boldsymbol{p}_i$ corresponds to a list of subtasks $[\boldsymbol{s}^i_1, ..., \boldsymbol{s}^i_{N_i}]$,  $\boldsymbol{s}^i_j$ is also a text string, indicating the $j$-th subtask for $i$-th milestone. This is achieved in the following steps. First, the LLM is prompted to extract hierarchical steps based on the subtitles of the video. Subsequently, the LLM is requested to modify the extracted steps in accordance with the user's query, as shown in Fig~\ref{sup_fig:plan}. Finally, we design a specific traversal algorithm that will only traverse the leaf nodes in order and send the corresponding subtask to the following modules.

\subsection{GUI Parser}
\begin{figure}[tbp]
\centering
\includegraphics[width=0.47\textwidth]{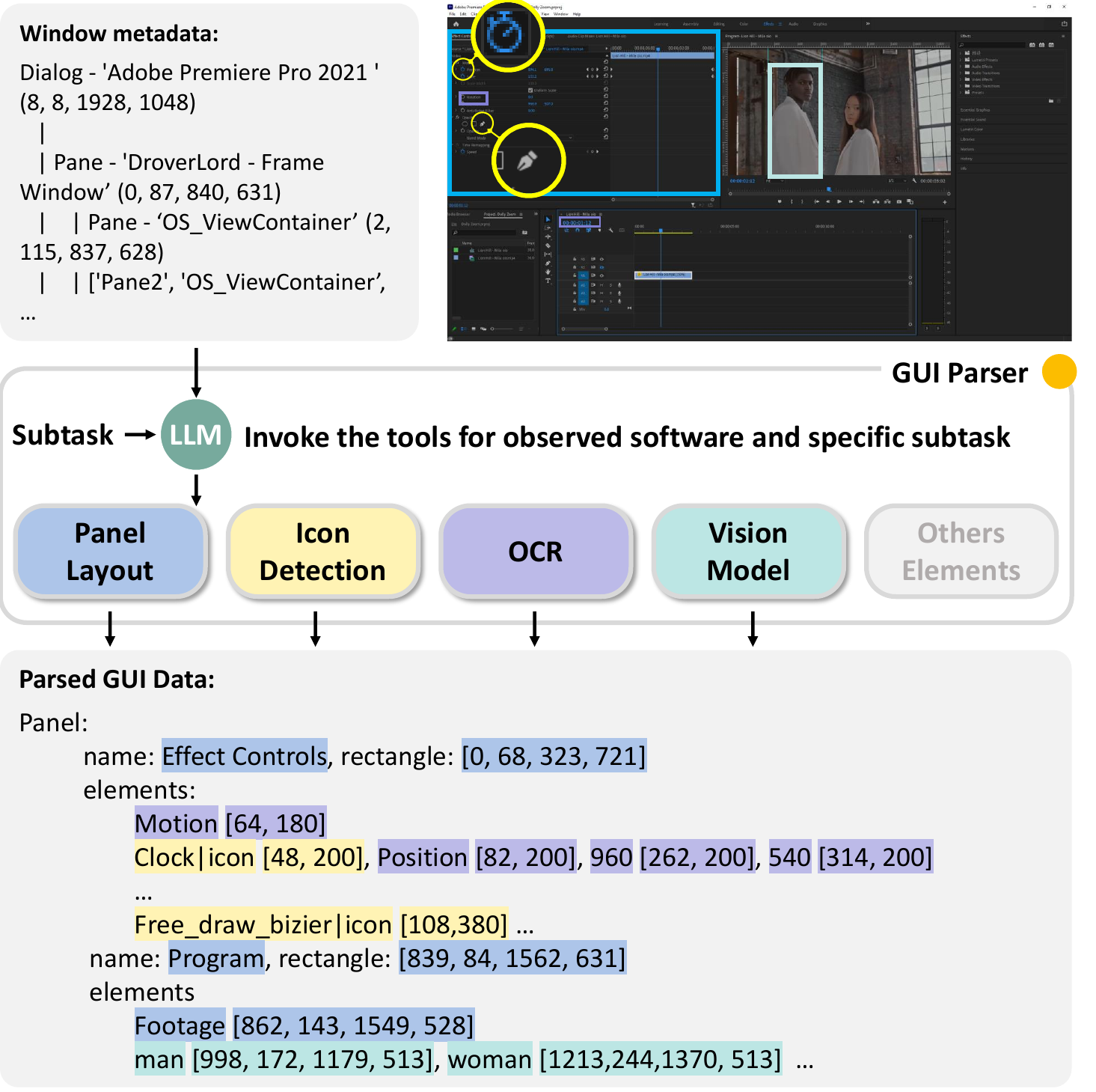} % Include the image
\caption{\textbf{Diagram Illustration of GUI Parser.} An LLM invokes different vision tools to parse various UI elements.} % Caption for the figure
\label{fig:gui} % Label for referencing the figure in the text
\vspace{-12pt}
\end{figure}

The goal of the GUI Parser is to convert an observed screenshot into a structured textual representation $\boldsymbol{o}_t$ like the Document Object Model (DOM). Given that desktop software typically comprises a wide variety of UI elements, it is hard for one model to extract all information, thus, we adopt approaches similar to MMReAct~\cite{yang2023mm} and VisualClues~\cite{xie2022visual}, invoking multiple tools to extract information, as shown in Figure~\ref{fig:gui}. Specifically, we utilize metadata from the system for panel segmentation, employ the OCR model to extract text from images and develop a pattern-matching method to identify icons. Additionally, some vision models, including a detector, and a segmentation model, are used to localize the objects in footage, and we have designed simple algorithms to extract specific elements such as scrolls and reference lines, etc. The GUI information is represented panel by panel, including the meanings of UI elements and their spatial position coordinates. 
% Figure~\ref{fig:vis} shows an example.

\subsection{Critic}
The Critic utilizes an LLM to evaluate the success of the executed action by analyzing the screenshots taken before and after the execution of the action $d(\boldsymbol{o}_t, \boldsymbol{o}_{t-1})$, where $d(.)$ is a function for identifying differences.
% for example, the Python DeepDiff API. 
It outputs four kinds of information: whether the previous action was executed correctly (a Boolean Success Flag), and if not, it provides an explanation; whether the current subtask is completed (Boolean Finish Flag), and if not, it offers an explanation, as shown in the Top of Figure~\ref{fig:ac}. The two flags and explanations, denoted as $\boldsymbol{c}_t$ will feed to the Actor.

\subsection{Actor}
\begin{figure}[tbp]
\centering
\includegraphics[width=0.45\textwidth]{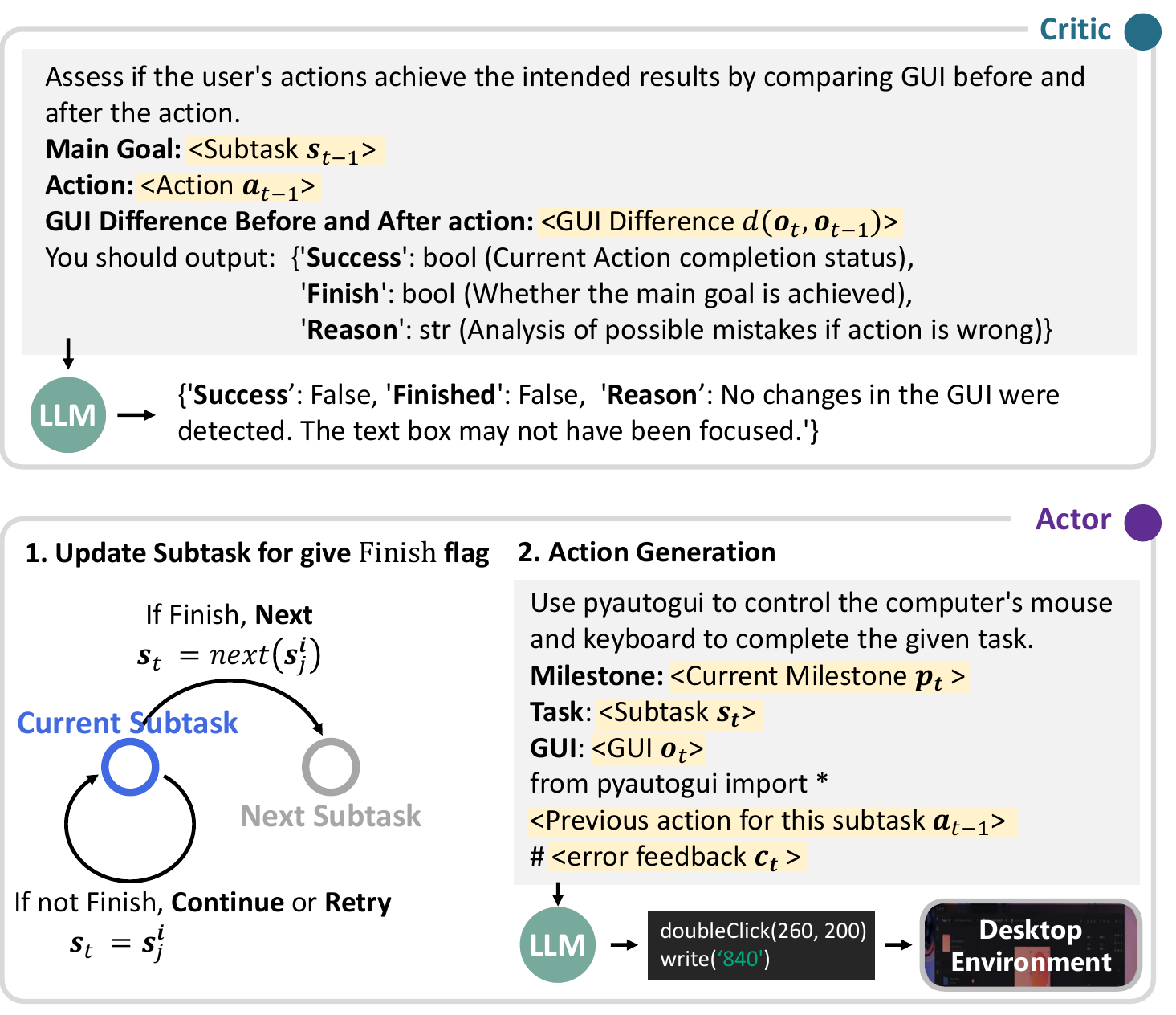} 
\vspace{-3pt}
\caption{\textbf{Top}: The \textbf{Critic} assesses the effectiveness of the previous action by analyzing the screenshots taken before and after its execution. \textbf{Bottom}: The \textbf{Actor} first updates the current subtask, then generates the subsequent action, considering the current observation, current subtask, historical actions, and Critic's feedback.} % Caption for the figure
\label{fig:ac} % Label for referencing the figure in the text
\vspace{-12pt}
\end{figure}

The Actor is built upon an LLM, aiming to generate actions within the action space of the \dataname benchmark. Specifically, given the Finish Flag provided by the Critic, the mode first plans what should be done next, as shown in Figure~\ref{fig:ac}. If the Finished Flag is False, the subtask $\boldsymbol{s}_t$ at time $t$ will still be $\boldsymbol{s}^i_j$, otherwise, $\boldsymbol{s}_t = next(\boldsymbol{s}^i_j)$, where $next(.)$ indicates moving to next subtask by using our designed traverse method illustrated in Sec.~\ref{sec:planner}.

Then, the Actor generates an output action by considering various factors: the current state of the observed software $\boldsymbol{o}_t$, the previous action $\boldsymbol{a}_t$, the current subtask $\boldsymbol{s}_t$, and its corresponding milestone $\boldsymbol{p}_t = parent(\boldsymbol{s}_t)$ (which indicates the milestone associated with the current subtask). Additionally, the Actor takes into account the Critic's feedback $\boldsymbol{c}_t$ on the performance of the previous action. 
Formally,
\begin{equation}
\boldsymbol{a}_{t} = \underset{\boldsymbol{a} \in \mathcal{A}}{\arg\max} \ p(\boldsymbol{a} | \boldsymbol{a}_{t-1}, \boldsymbol{o}_t, \boldsymbol{s}_t, \boldsymbol{p}_t, \boldsymbol{c}_t),
\end{equation}
where $\mathcal{A}$ denotes the action space, comprised of Python code. It's important to note that the output action $\boldsymbol{a}_{t}$ can either be a single action or a sequence of actions. This is implemented by prompting an LLM to process all the aforementioned information as input and subsequently generate the code for the next step, as illustrated at the bottom of Figure~\ref{fig:ac}.

\section{Experiments}
\textbf{Implementation Details.} In the following experiments, we use gpt-4-0613~\cite{openai2023gpt4} provided by OpenAI as the default LLM. In the GUI parser, we use Google OCR for extracting text, Yolo-v8~\cite{redmon2016you} to coarsely localize objects, and LangSAM~\cite{liu2023grounding,kirillov2023segment} to obtain the precise object contours. The difference module $d(.)$ is implemented by using DeepDiff~\cite{sep2023deepdiff}.

\subsection{Quantitative Results}
As \dataname is a novel Task that requires planning with an instructional video and processing Desktop environments (previous works mostly focus on Web or Android), there are no ready-made state-of-the-art models available for evaluation. Thus, we construct various variants based on our approach, to contrast some core concepts from previous works, thereby showing the effectiveness of our method and the challenges of \dataname.

\noindent\textbf{Comparision with SOTA Planning Method.}
In Table~\ref{exp:tab1}, we compare the planning approaches that have recently demonstrated exceptional performance in other environments. Specifically, we retained the GUI Parser and removed both the Planning and Actor-Critic modules. The subtitle of the instructional video is simply put into the prompt. Then, the model plans the steps in the following methods:
\begin{itemize}
    \item CoT~\cite{wei2022chain}: The CoT generates all the steps at once, which cannot obtain information from the environment.
    \item ReAct~\cite{yao2022react}: It iteratively interacts with the environment through a cycle of thought, action, and observation. 
\end{itemize}

\begin{table}[t]
    \centering
    \caption{Success rate (\%) of agents with different planning methods on \dataname. 
Human* represents the average performance of three non-expert humans who have viewed the instructional video only once, like how the model does. These results are a reference to better sense the extent of the model's capabilities.}
    \vspace{-0.3cm}
    \resizebox{0.43\textwidth}{!}{%
    \begin{tabular}{l|ccccc|c}
    \toprule
        Method & Design & Office & Widget & Sys. Set & File Mani. & Overall \\ \midrule
        CoT   & 5.9  & 10.8 & 20.0 & 0.00 & 36.7 & 12.0 \\ 
        ReAct & 14.7 & 27.0 & 50.0 & 62.5 & 63.6 & 32.0 \\ 
        \rowcolor{gray!25} Ours & 32.4 & 40.5 & 60.0 & 75.0 & 72.7 & 46.0 \\ \midrule
        Human$^*$ & 73.5 & 83.7 & 100.0 & 100.0 & 100.0 & 85.0 \\ \bottomrule
    \end{tabular}
    }
    \label{exp:tab1}
    % \vspace{-12pt}
\end{table}

The experimental results demonstrate that our model significantly surpasses previous planning methods. The result of CoT reveals that Desktop GUI Automation tasks often entail screen chanthus, thus, it is unable to cope effectively. Regarding ReAct, since it does not convert lengthy videos into discrete steps, it can operate on the finest granularity of action plans. Additionally, ReAct's absence of a dedicated module for evaluating and adjusting the planning path becomes a shortcoming, especially for complex tasks in office and design environments. The overall results indicate that \dataname poses significant challenges, especially for complex productivity tools. This difficulty arises from the intricacies involved in understanding and navigating sophisticated software interfaces, which require nuanced interpretation of visual elements and context-aware decision-making.

\noindent\textbf{Ablation on Planner, Actor and Critic.}
\begin{table}[t]
    \centering
    \caption{Success rate (\%) of agents with ablation of reasoning module.}
    % \vspace{-0.3cm}
    \resizebox{0.49\textwidth}{!}{%
    \begin{tabular}{l|ccccc|c}
    \toprule
        Method & Design & Office & Widget & Sys. Set. & File Mani. & Overall \\ \midrule
\rowcolor{gray!25} Full Model & 32.4 & 40.5 & 60.0 & 75.0 & 72.7 & 46.0 \\ 
        w/o Planning   & 20.6 & 27.0 & 50.0 & 75.0 & 63.6 & 35.0  \\ 
        w/o Critic     & 26.5 & 32.4 & 60.0 & 75.0 & 72.7 & 41.0 \\ 
        w/o Ins. Video & 11.8 & 37.8 & 60.0 & 62.5 & 72.7 & 37.0 \\ \bottomrule
    \end{tabular}
    }
    \label{exp:tab2}
    % \vspace{-14pt}
\end{table}
We also conducted ablation studies on our Critic and Planner, as shown in Table~\ref{exp:tab2}, where the w/o Planner method directly feeds the whole subtitle into Actor, instead of the parsed subtask. For simple tasks, the impact of these components was not particularly significant. However, their influence becomes much more apparent in complex Office and Design tasks. On another note, while the Critic appears to be a very important module, its performance enhancement was not as large as we initially expected. This is primarily because the Critic's judgments in complex tasks are not always accurate. It requires a high level of action-vision alignment, which still remains a relatively underexplored area, but we believe it is a direction worth exploring. Additionally, we constructed a variant that does not take into account the subtitles of videos. Instead, it utilizes GPT-4 to plan milestones and subtasks, denoted as w/o Ins. Video. This approach showed almost no significant performance loss in simple tasks because there weren't many alternative solutions. However, for the use of complex software like After Effects and Premiere Pro, instructional Videos proved to be very helpful.

\noindent\textbf{Ablation on GUI Parser.}
% \usepackage{booktabs}
% \usepackage{graphicx}
\begin{table}[]
\centering
\caption{Success rate (\%) of agents with ablation of GUI Parser.}
\label{tab:gui}
% \vspace{-0.40cm}
\resizebox{0.3\textwidth}{!}{%
\begin{tabular}{cccc|c}
\toprule
\multicolumn{4}{c|}{UI Elements}    & \multicolumn{1}{l}{} \\
Panel Layout & Icon & OCR & Others & Overall               \\ \midrule
\rowcolor{gray!25} \ding{51}         & \ding{51}     & \ding{51}    &  \ding{51}  &  46.0  \\ 
\color{red}{\ding{55}}         & \ding{51}     & \ding{51}    &  \ding{51}  &  44.0       \\
\ding{51}         & \color{red}{\ding{55}}     & \ding{51}    &  \ding{51}  &     13.0    \\
\ding{51}         & \ding{51}     & \color{red}{\ding{55}}    &  \ding{51}  &    4.0     \\ 
\ding{51}         & \ding{51}     & \ding{51}    &  \color{red}{\ding{55}}  &  43.0 \\ \midrule
\multicolumn{4}{c|}{Qwen-VL-Chat~\cite{bai2023qwen}}      &     5.0                 \\ 
\bottomrule
\end{tabular}%
}
% \vspace{-14pt}
\end{table}
Correctly parsing UI Elements is essential for generating actions. Here, we eliminate different UI elements in parsed GUI data to observe their impact. Table~\ref{tab:gui} shows that removing OCR had the most significant impact since text often contains crucial information in a GUI. Icons also led to notable performance loss, especially in Design and Office software, where many icons lack corresponding textual descriptions and are essential for specific functions. Interestingly, Panel Layout had minimal impact on performance, indicating GPT-4 can recognize the button without panel information, though it's still necessary for operations like clicking in a blank or margin of the panel area. The Others category, including footage content, scrolls, and similar elements, also had little effect. This is due to the model's current limitations in handling complex footage operations, even though they correctly recognize but still fail to complete the task. We also try to replace Qwen-VL-Chat~\cite{bai2023qwen} to replace the GUI Parser, allowing GPT-4 to plan button interactions and Qwen-VL-Chat to determine their positions. However, the results were not very satisfactory, as there may not be particular training for GUI button grounding.

\noindent\textbf{Impact of Large Language Model.}
% \usepackage{booktabs}
% \usepackage{multirow}
% \usepackage{graphicx}
We also experimented with different language models, gpt-3.5-turbo, and Llama2-7B~\cite{touvron2023llama}, in various modules, but found the results to be generally unsatisfactory, as shown in Table~\ref{tab:llm}. There are two main reasons for this: 1) The requirement for specific output formats. For instance, an action must be in the form of current step code and can only output code; any other content would render it non-executable. Similarly, the results from planning need to adhere to a certain format, which other language models sometimes fail to follow. 2) The issue of model hallucination. For the generation of actions, the model needs to stop at appropriate times, using updated GUI information to continue generating actions. However, non-GPT-4 models often hallucinate or invent too much information, leading to an incorrect code. For these relatively lightweight models to perform such customized functions effectively, they may require fine-tuning with specific datasets.

\subsection{Qualitative Results}
\begin{figure*}[bph]
\centering
\includegraphics[width=\textwidth]{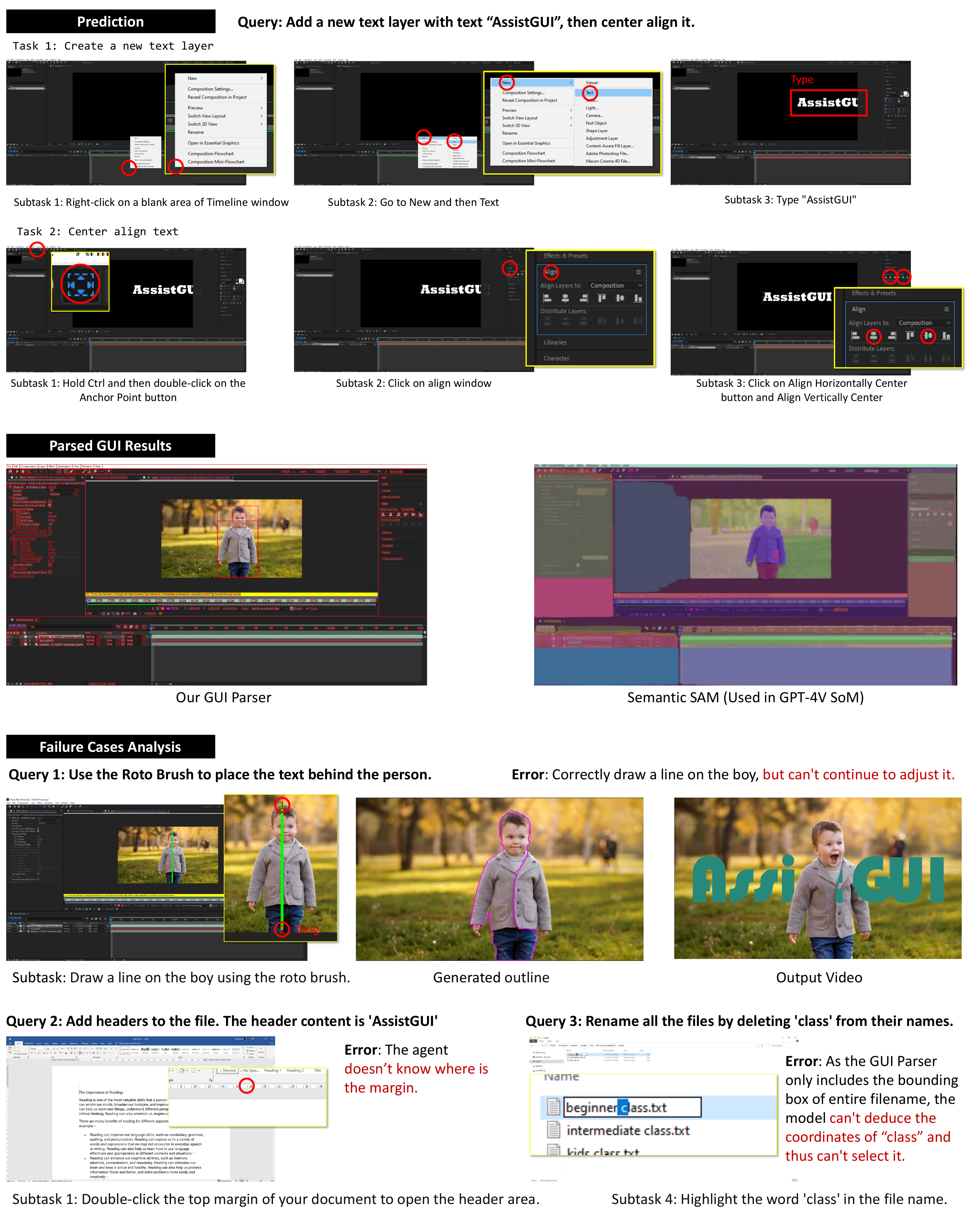} % Include the image
\caption{\textbf{Qualititave Results.} Top: We show one successful prediction. Middle: We compare our GUI Parser results with Semantic-SAM which is the core component for supporting GUI-4V to ground in Web or Smartphone Platform (i.e., GPT-4V-SoM). Bottom: We display some common errors with explanation.} % Caption for the figure
\label{fig:vis} % Label for referencing the figure in the text
\end{figure*}

In Figure~\ref{fig:vis}, we showcase some visualized results. Firstly, we present a successful prediction example, demonstrating that the model can effectively plan each step for relatively long processes, accurately perceive specific elements in the GUI, and convert them into the correct action code. Additionally, we display the performance of our designed Multi-modal LLM Agent, which can accurately identify most content, including small icons such as a clock-shaped keyframe button, checkboxes, and expand buttons. In contrast, although GPT-4V~\cite{openai2023gpt4v} possesses robust OCR capabilities, it fails to output button positions, rendering it unable to execute operations. The current best method to modify GPT-4V for button grounding is GPT-4V-SoM~\cite{yang2023setofmark}, which uses semantic-SAM to segment the image first, then label it, and finally input it to GPT-4V. This approach achieves remarkable results in Web and Android Navigation tasks. However, as seen, for desktop GUI understanding, the performance of GPT-4V-SoM is almost nullified due to the limitations of Semantic-SAM's segmentation capabilities in productivity software.

\begin{table}[]
\centering
\caption{Success rate (\%) of agents with ablation on LLM.}
\label{tab:llm}
% \vspace{-0.4cm}
\resizebox{0.31\textwidth}{!}{%
\begin{tabular}{cc|c}
\toprule
Planner & Actor \& Critic                  & Overall Score \\ \midrule
\rowcolor{gray!25}\multicolumn{2}{c|}{GPT-4}              &  46.0        \\ \midrule
\cellcolor{gray!25}              & GPT-3.5                 &  12.0\\
\multirow{-2}{*}{\cellcolor{gray!25}GPT-4}  & Llama2       &  1.0 \\
GPT-3.5 & \cellcolor{gray!25}                              &  19.0\\
Llama2  & \multirow{-2}{*}{\cellcolor{gray!25}GPT-4}       &  5.0    \\ \bottomrule
\end{tabular}
}
% \vspace{-16pt}
\end{table}

We also highlight some common errors encountered. 1) The model struggles with complex operations on footage, which can be highly intricate. For instance, Query 1 requires using a roto brush to select an object, necessitating continuous adjustments based on the generated edges, a capability our model currently lacks. Achieving this function might require training with specific samples or a more powerful Agent framework. 2) The model has difficulty understanding blurred areas, such as the edges of documents, blank spaces in Panels, or determining which area to select when multiple files are involved. 3) The spatial relationship in dense text. The granularity of OCR output bounding boxes is uncontrollable. Selecting a specific word or character in a text segment is not straightforward with the current OCR predictions. This may require a highly versatile text grounding model to address effectively.
% We show some 

\begin{figure}[tbp]
\centering
\includegraphics[width=0.48\textwidth]{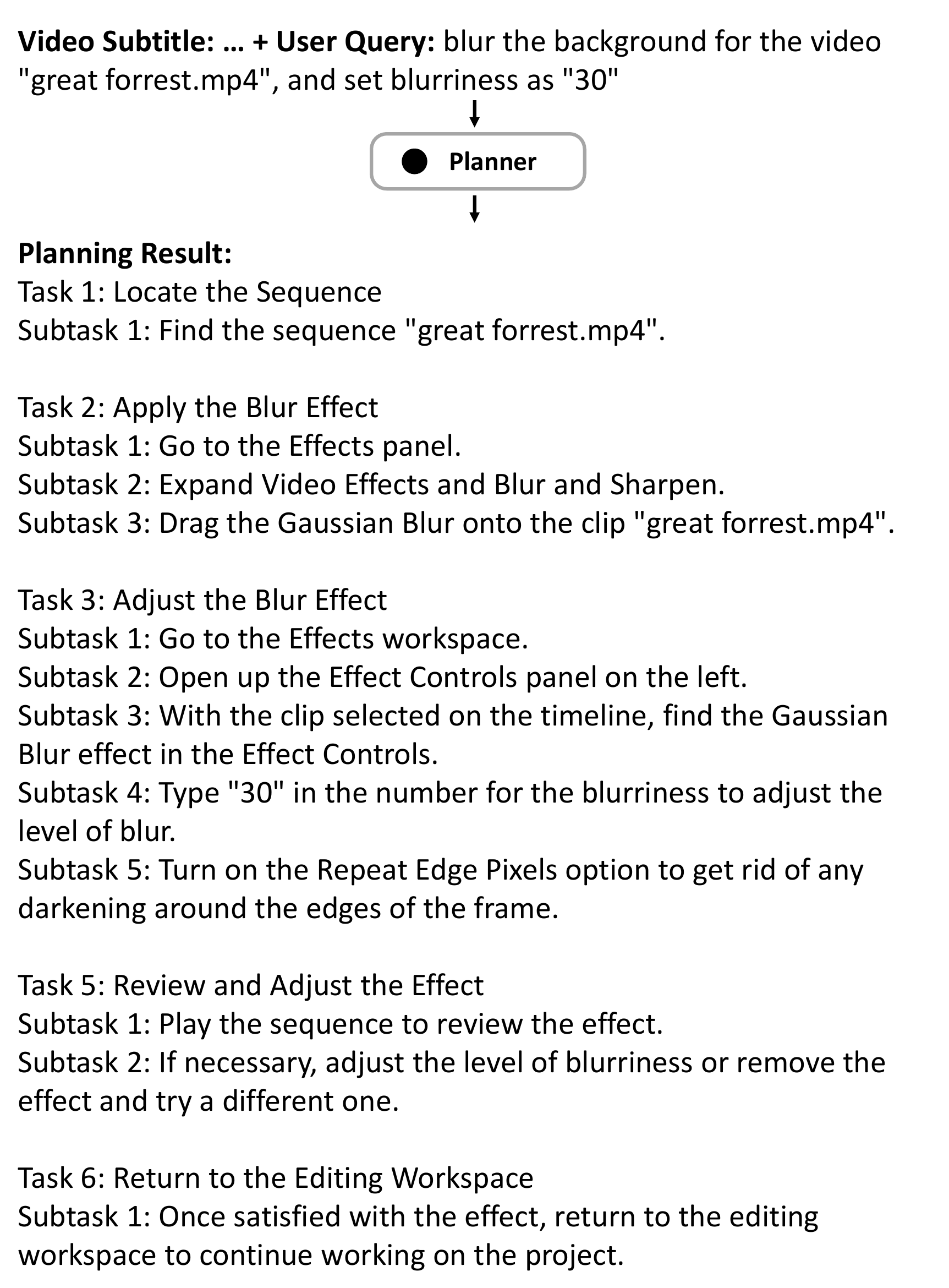} % Include the image
\vspace{-0.3cm}
\caption{\textbf{Planning Results.} The UI elements are organized panel by panel.} % Caption for the figure
\label{sup_fig:plan_result} % Label for referencing the figure in the text
\end{figure}

In Figure~\ref{sup_fig:plan_result}, we present an example of the \textbf{Planner} prediction. The results show that, despite the strength of GPT-4, the predictions still have some flaws, such as including redundant operations. For instance, Task 6 does not actually correspond to any specific action. This issue mainly arises from the fact that these steps are included in the instructional video, and GPT-4 cannot definitively determine whether to exclude them.

\begin{figure}[tbp]
\centering
\includegraphics[width=0.45\textwidth]{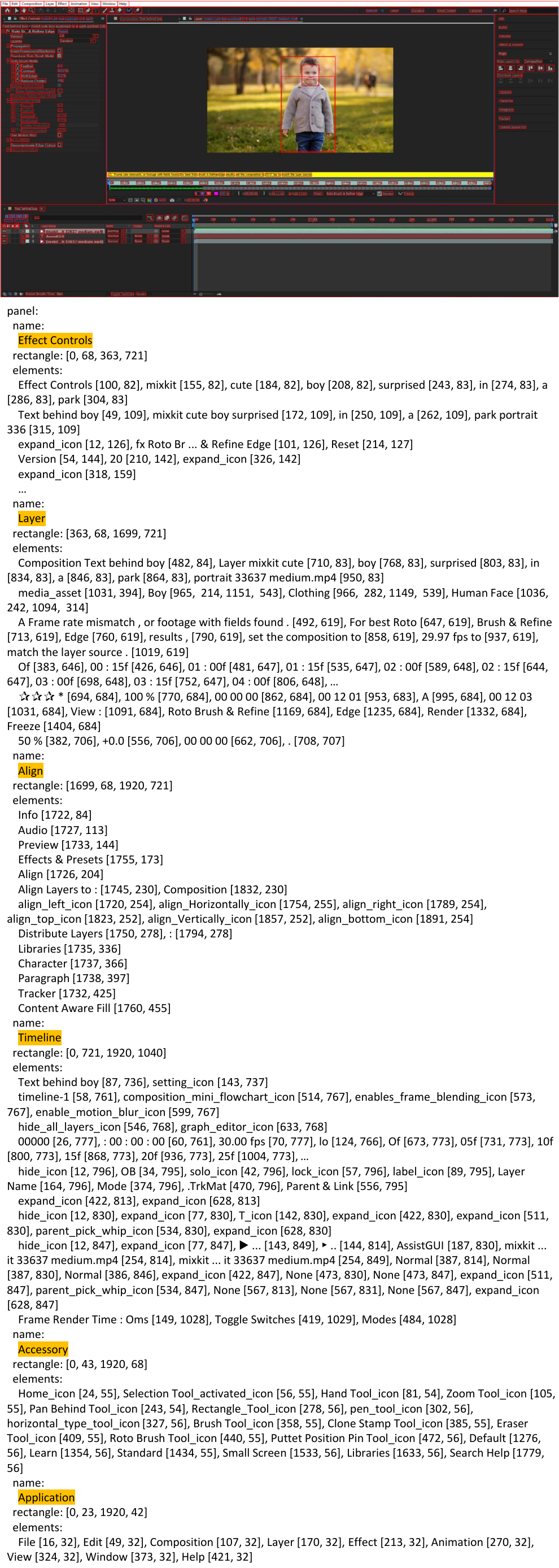} % Include the image
\vspace{-0.3cm}
\caption{\textbf{Parsed GUI Results.} The UI elements are organized panel by panel.} % Caption for the figure
\label{sup_fig:gui_result} % Label for referencing the figure in the text
\end{figure}

\begin{figure*}[t]
\centering
\includegraphics[width=0.98\textwidth]{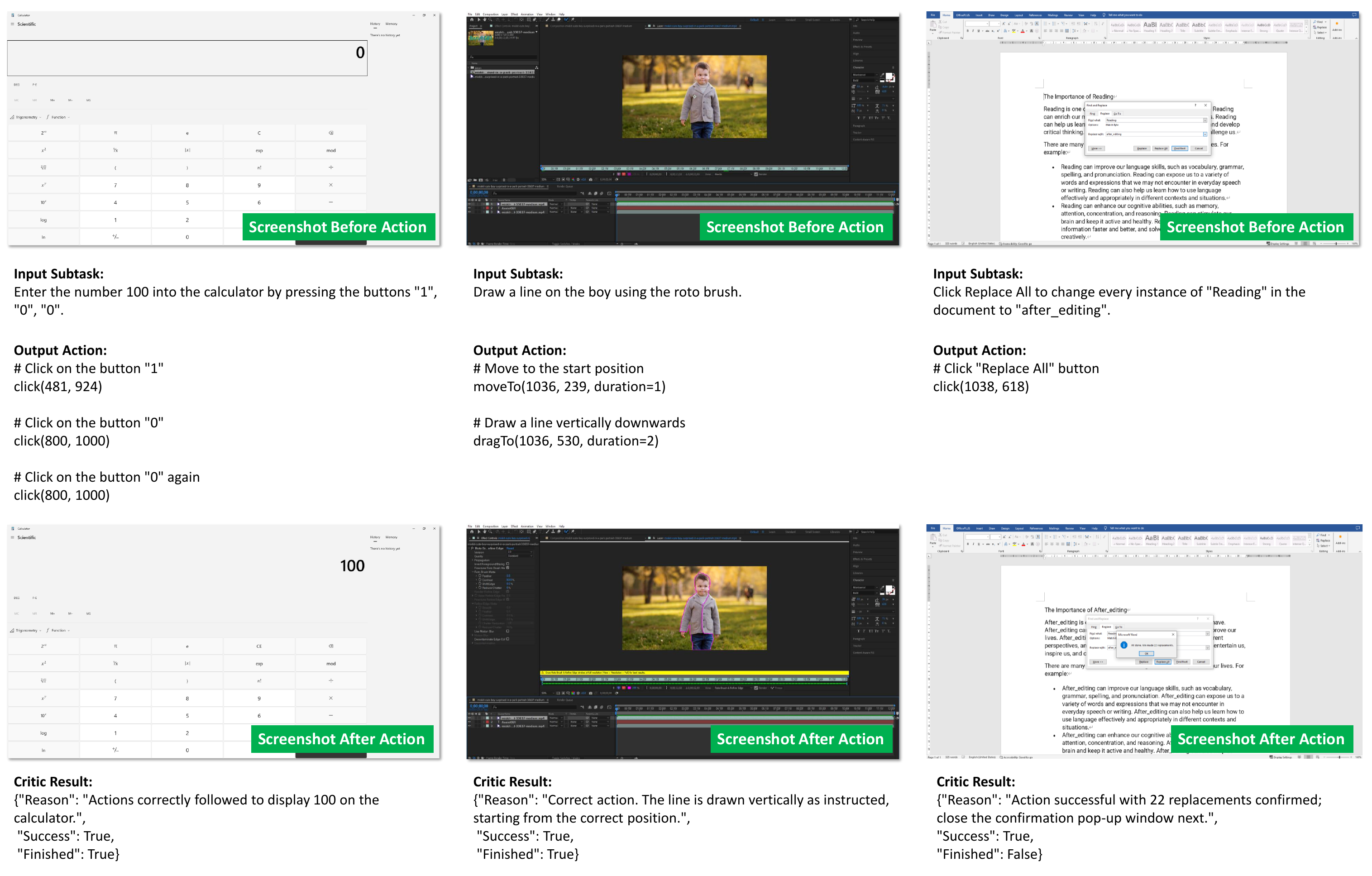} % Include the image
\vspace{-0.3cm}
\caption{\textbf{Prediction Results of Actor and Critic Module.} We show the prediction results of one specific subtask in solving a query.} % Caption for the figure
\label{sup_fig:ac_result} % Label for referencing the figure in the text
\end{figure*}

In Figure~\ref{sup_fig:gui_result}, we show one example of the outputs of \textbf{GUI Parser}. The model can detect most UI elements, but there are still some flaws. 1) There are still errors in text detection. For example, there are some issues with the detection of timestamps. The timestamp in the lower-left corner should be 0:00:00:00, but it is detected as : 00 : 00 : 00. The numbers in the Timeline in the lower-left corner are not detected. 2) Some visual elements are still difficult to recognize, such as the long bars on the timeline corresponding to each layer. Additionally, the current method is unable to understand some curves and figures, and it might be necessary to leverage the capabilities of GPT-4V in the future.

In Figure~\ref{sup_fig:ac_result} we showcase prediction examples from the \textbf{Actor and Critic} modules. It is evident that the model is capable of not only producing individual step actions but also generating a continuous action sequence. Additionally, for the Critic module, current models can effectively judge the outcomes of some simple actions, such as clicking action, as demonstrated in the left and right examples. However, for more complex scenarios, such as determining whether an object has been completely cropped out, as seen in the middle case, the model still lacks the capability to perceive this accurately.
\section{Comparison with Previous Benchmarks}
\label{supp_sec:rel}
We discuss the differences between our approach and existing benchmarks in the following aspects, as shown in Table~\ref{tab:benchmarks}:

\textbf{Platform:} Previous methods~\cite{toyama2021androidenv,yao2022webshop,wen2023empowering} mainly focused on Web and SmartPhone platforms, such as AndroidEnv, AutoDroid, and WebShop. AssistGUI, however, concentrates on desktop operations. This distinction primarily brings about differences in GUI complexity. The complexity on desktops is significantly higher than on other platforms, mainly reflected in the density of information, the diversity of visual elements, and the diversity of panel layouts.

\textbf{Task Focus:} Exisiting methods~\cite{toyama2021androidenv,yao2022webshop,wen2023empowering} primarily study two types of tasks. One category is games, for instance, the majority of tasks in AndroidEnv are games, such as FlappyDroid, and Pong. The characteristic of game tasks is that the environment has a clear reward, making it easy to measure the performance of the model. Additionally, for most games, the types of operations are relatively limited. The other category includes web navigation and basic smartphone operations. These tasks have relatively simple operational patterns. For example, web navigation mainly involves buying a series of items according to requirements, with the difficulty lying in planning what to buy. The operations are relatively limited in type.

The distinguishing feature of \dataname is its focus on the use of productivity tools. The challenge of this category of tasks lies in the possibility of encountering new types of operations with different software. For instance, with After Effects, one might need to perform some drawing on the material. This presents a more formidable challenge for the model's understanding of the GUI and the generation of actions.

\textbf{Dataset Scale and Annotation:} Previous benchmarks~\cite{toyama2021androidenv,yao2022webshop,wen2023empowering} mainly involved about a hundred tasks. WebShop is somewhat unique; it primarily consists of one task, which is purchasing items, but it comes with different instructions specifying various purchasing requirements. The dataset scale of our benchmark is similar. However, a distinctive feature of our tasks is the use of professional software to modify documents or materials. Therefore, we also provide some project files to ensure that all methods start from the same initial state.

\begin{table*}[t]
    \centering
    \caption{\textbf{Comparison of related benchmarks.} \dataname is unique in its platform and task focus. It additionally provides project files for each task.}
    \resizebox{0.8\textwidth}{!}{%
    \begin{tabular}{l|ccccc}
    \toprule
        Benchmark & \# APPs & \# Tasks & Platform & Task Focus & Project File \\
        \midrule
        AndroidEnv~\cite{toyama2021androidenv} & \textasciitilde30 & >100 & Android OS & Game \& App Usage & \ding{55}\\
        WebShop~\cite{yao2022webshop} & 1 & 1 task, 12K instructions & OpenAI Gym & Web-based e-commerce & \ding{55} \\
        AutoDroid~\cite{wen2023empowering} & 13 & 158 & Android OS & App Usage & \ding{55} \\ \midrule
        AssistGUI & 9 & 100 & Windows & Productivity Software Usage & \ding{51}  \\ 
        \bottomrule
    \end{tabular}
    }
    \label{tab:benchmarks}
\end{table*}

\section {Conclusion}

This paper introduced \dataname, a novel benchmark for assessing the capability of models to manipulate the mouse and keyboard on the Windows platform in response to user requests. To this end, we collected a diverse set of 100 tasks across 9 widely-used applications, ensuring each task was supplemented with the necessary project files for a fair evaluation. We also presented our Actor-Critic Embodied Agent framework, a significant step forward in the realm of GUI automation. This framework is anchored by a GUI parser driven by an LLM-agent, coupled with an enhanced reasoning mechanism. This design is particularly adept at handling complex, lengthy procedural tasks that are commonplace in professional software environments. Our experimental results were promising, demonstrating that our approach notably outperforms existing methods in GUI automation. However, despite these advancements, our findings also highlight the considerable challenges that remain in this field. 

{
    \small
    \bibliographystyle{ieeenat_fullname}
    \bibliography{main}
}


\end{document}